\pdfoutput=1

\documentclass[11pt]{article}

\usepackage[preprint]{acl}

\usepackage{times}
\usepackage{latexsym}

\usepackage[T1]{fontenc}

\usepackage[utf8]{inputenc}

\usepackage{microtype}

\usepackage{inconsolata}

\usepackage{graphicx}
\usepackage{booktabs}
\usepackage{array}
\usepackage{arydshln}
\usepackage{paralist}
\usepackage[inline]{enumitem}
 
\title{Exploring ReAct Prompting for Task-Oriented Dialogue: Insights and Shortcomings}

\author{
  \textbf{Michelle Elizabeth\textsuperscript{1,2,3}},
   \textbf{Morgan Veyret\textsuperscript{3}},
  \textbf{Miguel Couceiro\textsuperscript{1}},\\
  \textbf{Ond\v{r}ej Du\v{s}ek\textsuperscript{2}},
  \textbf{Lina M. Rojas-Barahona\textsuperscript{3}},
\\
  \textsuperscript{1}University of Lorraine/LORIA, France,
  \textsuperscript{2}Charles University, Czechia,
  \\
  \textsuperscript{3}Orange Innovation, France
\\
  \small{\href{mailto:m@domain}{michelle.elizabeth@orange.com},}
  \small{\href{mailto:m@domain}{morgan.veyret@orange.com},}
  \small{\href{mailto:m@domain}{miguel.couceiro@loria.fr},}
  \\
  \small{\href{mailto:m@domain}{odusek@ufal.mff.cuni.cz},}
  \small{\href{mailto:m@domain}{lina.rojas@orange.com},}
}

\begin{document}
\maketitle
\begin{abstract}
Large language models (LLMs) gained immense popularity due to their impressive capabilities in unstructured conversations. 
Empowering LLMs with advanced prompting strategies such as reasoning and acting (ReAct)~\cite{yao2022react} has shown promise in solving complex tasks traditionally requiring reinforcement learning. In this work, we apply the ReAct strategy to guide LLMs performing task-oriented dialogue (TOD). We evaluate ReAct-based LLMs (ReAct-LLMs) both in simulation and with real users. While ReAct-LLMs severely underperform state-of-the-art approaches on success rate in simulation, this difference becomes less pronounced in human evaluation. Moreover, compared to the baseline, humans report higher subjective satisfaction with ReAct-LLM despite its lower success rate, most likely thanks to its natural and confidently phrased responses.
\end{abstract}

\section{Introduction}

Task-oriented Dialogue (TOD) systems solve 
tasks, such as accessing information or booking places and tickets, by interacting  with humans in natural language \cite{budzianowski-etal-2018-multiwoz,rastogi2020towards}. 
Traditional TOD pipelines integrate specialized components for natural language understanding (NLU), dialogue state tracking (DST), dialogue management and natural language generation (NLG), and optionally speech recognition and synthesis \cite{ultes2017pydial}.
On the other hand, end-to-end architectures model all components in a single neural networks~\cite{wen-etal-2017-network, zhu2020convlab, zhu2022convlab3}. Both approaches are costly to develop, requiring manual engineering and/or collecting large dialogue corpora. Large language models (LLMs) \cite{ouyang2022traininglanguagemodelsfollow} offer an alternative to this by generalizing from instructions or a small number of examples and promise fluent and natural replies. However, unlike standard LLM use scenarios, task-oriented dialogues typically have a rigid structure and require access to an external database to retrieve necessary information, such as venues or objects to search for and their properties. 

Recently, synergizing reasoning and acting in LLMs (ReAct)~\citep{yao2022react} has shown promising results in controlled tasks that need external information access. ReAct employs few-shot LLM prompting with a sequence of \emph{thoughts}, \emph{actions}, and \emph{observations}. Thoughts refer to internal reasoning that decomposes a problem into sub-problems. Actions execute external API calls or programs, and observations analyze the results of actions. 
In this work, we investigate the ability of LLMs guided by ReAct to solve task-oriented dialogue. 

Our contributions are as follows:
\begin{enumerate*}[label=(\roman*)]
\item We implement a ReAct-based system for TOD,  using GPT-3.5 and GPT-4 LLMs respectively~\cite{openai2024gpt4technicalreport}, for the MultiWOZ tourist information domains~\cite{budzianowski-etal-2018-multiwoz}.
\item  We evaluate our ReAct-LLM system both with a simulated user and with humans, and we include a manual qualitative analysis of human dialogues and a API cost versus performance trade-off assessment.
\item  Our results show that ReAct-LLMs clearly underperform state-of-the-art baselines in terms of success rate in simulation. However, this difference is less pronounced in human evaluation, and humans even rate their satisfaction with the ReAct-LLM system higher than the baseline.
\end{enumerate*}

\section{Related Work}

Classical approaches for dialogue management \cite{casanueva2018benchmarkingenvironmentreinforcementlearning,weisz2018sample}, varied from handcrafted (HDC) to reinforcement learning (RL). The combination of deep RL with imitation learning~\cite{cordier2020dilutednearoptimalexpertdemonstrations} as well as structural RL have also been applied to multi-domain, multi-task dialogue~\cite{chen2018structured, cordier2022graphneuralnetworkpolicies}. 
However, these approaches require separate specialized components, involving extensive engineering, the need for semantically annotated data as well as user simulators operating at the semantic level.

Latest end-to-end neural architectures are built on top of pretrained language models~\cite{peng_soloist_2021,lin_mintl_2020} and involve two-step generation: the model first generates the dialogue state or database query based on user input; then, it generates the reply based on external database search.
These approaches provide more flexibility and potentially better fluency, but require even larger training corpora.

Recent approaches explore simple zero-shot or few-shot LLM prompting for TOD, using examples of relevant dialogue turns \cite{hudecek-dusek-2023-large}, database or API definitions \cite{stricker_few-shot_2024}, templated conversation snippets \cite{zhang_sgp-tod_2023}, or detailed behavior and API calling instructions \cite{xu-etal-2024-rethinking}. 
In contrast, in this work we propose to use the ReAct strategy \cite{yao2022react} to guide LLMs towards task oriented dialogue through domain selection, slot detection and belief state tracking.

Since dialogue is dynamic 
and evaluation on static data in single-turn replies may not be consistent with full dialogue performance \cite{takanobu_is_2020},
we evaluate the system on full dialogues. First, we evaluate performance in a simulated environment, which, unlike~\citet{xu-etal-2024-rethinking}, utilizes the well established \textsc{ConvLab3} benchmark \cite{zhu2022convlab3} to guarantee reproducibility and fair comparison with previous approaches. We also assess the conversations with humans.
Unlike previous works on dialogue management~\cite{10.1109/TASLP.2018.2851664,zhu2020convlab,cordier2022graphneuralnetworkpolicies}, 
the simulator and our system do not interact at the semantic level, instead they interact in natural language. 
Unlike previous end-to-end and LLM-based approaches, which generated delexicalized responses with placeholders and replaced these with valid entity names from the database in a rule-based fashion, we generate full responses including entity names.
This avoids constraining the inherent capabilities of LLMs in generating natural language. Instead, we guide the LLM towards the database constraints through ReAct prompting.


\section{ReAct for Task-Oriented Dialogue}

\begin{figure*}[t]
    \centering
    \includegraphics[width=0.75\textwidth]{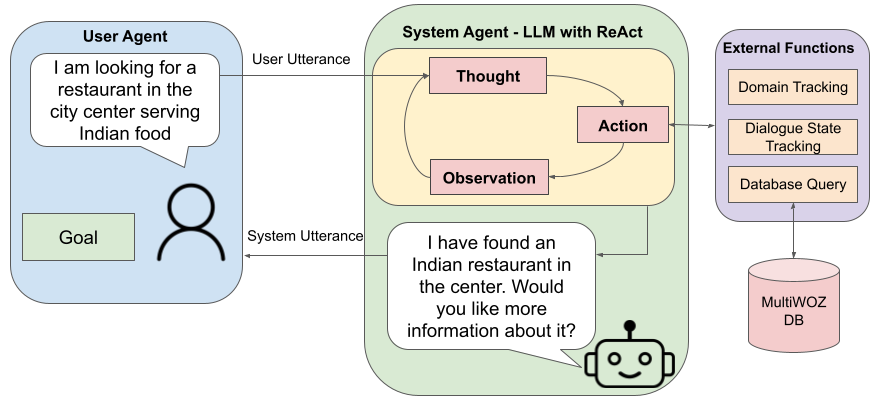}
    \caption{The proposed ReAct-LLM system agent uses few-shot examples in the prompt to guide the LLM in decomposing reasoning into a sequence of thoughts, actions, and observations.}
    \label{fig:react-tod-setup}
\end{figure*}

Figure~\ref{fig:react-tod-setup} shows the proposed architecture: the ReAct-LLM system agent (SA) interacts with a user in natural language. The SA has access to external tools to guide it through the TOD pipeline.
We provide few-shot examples in the prompt following ReAct \cite{yao2022react}. We give the ReAct-LLM agent a list of tools: \texttt{list\_domains}, \texttt{list\_slots}, \texttt{db\_query} and \texttt{generate\_booking\_reference} (details in Appendix \ref{app:tool}), which it can use as actions. The example provided in the prompt shows how and in what order the tools should be used. The reasoning process to be followed by the SA is outlined below:

\medskip
\noindent
\textbf{Step 1:} The SA should try to understand the user input. Its first thought should be explicitly planning out the next course of action, to be executed in Steps 2-5. The SA should plan for identifying the domain and the corresponding slots and values from the user input, forming the belief state and using it to query the database, retrieving the results and forming the final answer based on the results.
   
\noindent
\textbf{Step 2:} The SA should call \texttt{list\_domains}, observe its output and decide which domain the user request belongs to.

    
\noindent
\textbf{Step 3:} Now it needs the list of slots available in the selected domain to identify the slot values from the user request. The system should use the \texttt{list\_slots} tool with the domain as input, observe the list of slots available and decide which slots have been mentioned in the user input and form/update \textit{the belief state}.


\noindent
\textbf{Step 4:} It should call \texttt{db\_query} using \textit{the state} as input to retrieve records from the database that match the user request.

\noindent
\textbf{Step 5:} As the final step, it observes the retrieved entities and generates an appropriate response.

We also provide a tool for generating booking reference numbers when the user requests a booking (\texttt{generate\_booking\_reference}). The agent is given a detailed example in the prompt showing the sequence of thoughts, actions and observations for two consecutive user turns (see Appendix \ref{app:prompt}). It also has access to previous conversation history as well as a description of each tool provided.

\section{Experimental Setup}
We use the LangChain\footnote{\url{https://www.langchain.com/}}  library  for implementing ReAct-LLM. The prompt details are given in Appendix~\ref{app:prompt}, Figure~\ref{fig:react-prompt}. We use Langfuse\footnote{\url{https://langfuse.com/}} for debugging the reasoning traces and to keep track of the computational costs of our experiments. We experiment with OpenAI GPT-3.5 (\emph{gpt-3.5-turbo-0301}) and GPT-4 (\emph{gpt-4-32k}) models. 
We first couple our SA with a simulated user, then proceed to evaluate it with humans.

\subsection{Simulated User}

We implemented an LLM-based user agent, but it was not able to end the conversation correctly and occasionally switched its role to play the system, which resulted in inappropriate conversations (see Appendix~\ref{a:llmuser}). Therefore, we use the agenda-based user simulator~\cite{schatzmann-etal-2007-agenda} implementation in \textsc{ConvLab 3}~\citep{zhu2022convlab3}.
A goal generator in agreement with the MultiWOZ dataset is used to initialize the agenda. The simulator then generates the semantic representation 
and converts it into natural language. This is fed to the system agent and the response from the system is sent to the BERT-based NLU of the simulator that returns its semantic representation. This semantic representation is in turn used to update the agenda. The simulator then generates the next utterance based on the system response and the updated agenda.

\subsection{Evaluation Setup}
\label{sec:exp-setup}

To measure how well the user goals were satisfied by the system, we compute the standard metrics: \textit{success}, \textit{book}, \textit{inform} and \textit{complete} rates as well as \textit{the average number of turns},\footnote{Note that the lower the average number of turns the better since this means that the system achieves the goal faster.} using \textsc{ConvLab 3}~\citep{zhu2022convlab3}. A dialogue is successful if the system provided the right information and was able to book the requested entities in agreement with the user goal. 
In addition, we also measure the cost incurred by using OpenAI APIs.
We compare our ReAct-LLMs to multiple dialogue management baselines: HDC and RL-based, such as proximal policy optimization (PPO) \cite{schulman_proximal_2017} and structured  RL with imitation learning (ACGOS)~\cite{cordier2022graphneuralnetworkpolicies}. These baselines use \textsc{ConvLab}'s pipelines with \textsc{BERT NLU} and template-based NLG. 
We simulated 1000 dialogues, using a fixed random seed in the goal generator for reproducibility.

For a more realistic assessment, we decided to invite humans to evaluate the system for a fair assessment.
We opt for an in-house evaluation to ensure high quality.
Finally, we do a detailed qualitative analysis of a small sample of the dialogue logs to identify the most frequent error sources.
%

\section{Results}
We now assess both simulated and real users.
\begin{table*}[ht!]
\centering
\resizebox{.9\textwidth}{!}{%
\small
\setlength{\tabcolsep}{5pt}
\begin{tabular}{l>{\hspace{-12pt}}ccccc}
\toprule
\textbf{Configuration }         & \textbf{Avg Turns $\downarrow$}                   & \textbf{\begin{tabular}[c]{@{}c@{}}Inform  Rate $\uparrow$\\ (P/R/F1)\end{tabular}} & \textbf{Book Rate $\uparrow$}                   & \textbf{Success Rate $\uparrow$} & \textbf{Complete Rate $\uparrow$}               \\
\midrule

HDC (semantic level)           & {\color[HTML]{E97132} \textbf{10.6}} & 87.2 / 98.6 / 90.9                                                        & 98.6                                 & {\color[HTML]{E97132}\textbf{97.3}}         & 97.9                                 \\

ACGOS  (semantic level)  & 13.2                                 & {\color[HTML]{E97132} \textbf{94.8 / 99.0 / 96.1}}                              & {\color[HTML]{E97132} \textbf{98.7}} & 97.0                  & {\color[HTML]{E97132} \textbf{98.2}} \\
\hline

BERT NLU + HDC + Tpl.                                                                           & \textbf{12}                                   & 82.8 / 94.1 / 86.2                                                        & \textbf{91.5}                        & \textbf{83.8}         & \textbf{92.7}                        \\

BERT NLU + PPO + Tpl.                                                                           & 17.8                                 & 69.4 / 85.8 / 74.1                                                        & 86.6                                 & 71.7                  & 75.5                                 \\

BERT NLU + ACGOS + Tpl.                                                                & 14.8                                 & \textbf{88.8 / 92.6 / 89.5}                                               & 86.6                                 & 81.7                  & 89.1                                 \\
\hline

\textbf{ReAct-LLM (GPT-3.5)}                                                             & 15.3                                 & 59.0 / 64.9 / 58.3                                                          & 40.5                                 & 28.2                  & 45.9                                 \\

\textbf{ReAct-LLM (GPT-4)}                                                               & 15.5                                 & 62.7 / 81.3 / 66.8                                                          & 58.2                                 & 43.6                  & 63.8    \\
\bottomrule
\end{tabular}%
}
\caption{Simulated evaluation on 1000 dialogues (Section~\ref{sec:exp-setup}). Tpl. stands for templated-based NLG.} 
\label{tab:e2e-metrics}
\end{table*}


\subsection{Simulated Evaluation}

Table \ref{tab:e2e-metrics} compares ReAct-LLM systems with previous works in \textsc{ConvLab}. 
Note that systems in the first section of the table interact at the semantic level and hence, the metrics show the upper bound of the performance that can be achieved by a full TOD system.
We observe that ACGOS performs best in terms of \textit{inform rate}. The HDC policy proves to be superior to other methods with higher \textit{book, success and complete rates} and fewer \textit{turns}.  Both React-LLMs clearly underperform all baselines, with the GPT-4 version distinctly ahead of the GPT-3.5 one.

ReAct-LLMs are costly. Table \ref{tab:gpt-cost} shows the per-token cost for the two GPT models, as well as the total number of tokens used and the approximate total experiment cost. 
The success rate improvement for GPT-4 does not appear to justify the cost difference.

\begin{table}
\centering\small
\setlength{\tabcolsep}{3pt}
\begin{tabular}{lrrrr}
\toprule
& \multicolumn{2}{c}{\bf Cost/1M tokens} & \multicolumn{2}{c}{\bf 1000 sim.~dialogues} \\
\textbf{Model} & \textbf{Input} & \textbf{Output} & \bf \# Tokens & \bf Total cost \\
\midrule 
gpt-3.5-turbo-0301\hspace{-3pt} & \$1.50  & \$2.00 & 40.6M & \$61.71 \\
gpt-4-32k          & \$60.00 & \$120.00  & 35.8M & \$2,258.81 \\
\bottomrule
\end{tabular}%
\caption{Experiment cost for GPT-3.5 vs.~GPT-4.}
\label{tab:gpt-cost}
\end{table}

\subsection{Human Evaluation}
Volunteers were asked to chat online with a system, in order to achieve a dialogue goal randomly generated by ConvLab. They could start a conversation as many times as they wish and were paired with a randomly chosen system each time, either the HDC baseline or our ReAct-LLM (GPT-3.5). They were asked to give a score for success (if all the dialogue goals were achieved) as well as a subjective score for their overall satisfaction with the conversation on a scale of 1-5. We then converted the ratings to percentages and reported the average rating provided for each system.
We collected 95 dialogues for each system, with the help of 20 users. Table~\ref{tab:humantableconvlab} shows that HDC performs better in terms of \textit{success rate}, but not by the same margin it had in simulation. 
We see that the HDC system falls short in the human evaluation compared to the user simulation. Contrary to the simulated evaluation, ReAct-LLM performs much better with real users. Overall, users are more satisfied with ReAct-LLM than with HDC, despite the better success rate of HDC, because ReAct-LLM is self-confident and generates well structured, polite, fluent and natural sentences even though it was not always able to find what the user was looking for. We also see that the React-LLM system has a slightly lower average number of turns when compared with the simulated evaluation, while the opposite is true for HDC.

\begin{table}[!tb]
\centering
\small
\setlength{\tabcolsep}{3pt}
\begin{tabular}{lcccc}
    \toprule
    \textbf{Dialogue System} & \textbf{Avg} & \textbf{Satisfaction} & \textbf{Success}\\
     & \textbf{Turn} & \textbf{Rate (\%)} & \textbf{Rate (\%)}\\
    \midrule
    BERT NLU + HDC + Tpl. & 15.91 & 54.10 & \textbf{60.00}\\
    \textbf{ReAct-LLM (GPT-3.5)} & \textbf{14.42} & \textbf{65.47} & 50.52\\

    \bottomrule
\end{tabular}
\caption{Human evaluation results, with 95 dialogues for each system.}
\label{tab:humantableconvlab}
\end{table}

\subsection{Qualitative Analysis}

By inspecting a sample of the generated dialogues, we identify several issues.
First, we see that the reasoning traces may just be imitating the examples given in the prompt. This may work for simpler cases with fewer goals to achieve. However, when the goals get larger with multiple domains and the user requests become more complicated, ReAct-LLM struggles to understand the user and to perform tasks accordingly. Beyond that, the reasoning is inconsistent and strays from instructions at times.
Furthermore, the LLM can come up with creative responses, but struggles to stay within the bounds set by the instructions, often producing invalid dialogue states or not sticking to the set of external tools given.
Compared to the GPT-3.5 agent, the GPT-4 agent is more consistent with respect to the desired reply format, is better at clarifying, and produces more verbose and polite replies.
More details are shown in Appendix~\ref{app:qualit} and~\ref{app:gpt35vs4}.

\section{Conclusion}

The performance of ReAct-LLM falls short compared to HDC and RL baselines, which perform better mainly due to their fine-grained control at each step in the pipeline. By relying completely on the reasoning abilities of ReAct-LLM, we lose the ability to control its reasoning traces and response generation. Additionally, difficulty in understanding the system requests by the user simulator, due to \textsc{BERT-NLU} errors, leads to repeated utterances and thus a higher number of turns on average. The human evaluation shows that ReAct-LLM is preferred by users over the HDC baseline, despite its lower success rate.

\bibliography{custom}

\appendix

\section{The ReAct Prompt}
\label{app:prompt}

Figure~\ref{fig:react-prompt} shows an example of the final ReAct prompt, namely Generic Prompt, in which the examples provided in the prompt (Figure \ref{fig:react-example}) contained a random example from the MultiWOZ dataset. We also experimented with another variation (i.e., Domain Specific), in which the examples provided were dynamically changed based on the domains in the goal. We observe that using domain-specific examples in the prompt has no effect on the performance of the system. Our results (see Table~\ref{tab:prompt-metrics}) show that the system in fact performs slightly better when there is only one random example irrespective of the domains of the user goal.
\subsection{Tool Description}
\label{app:tool}
The tools are python functions that we provide in the prompt to guide the model to make grounded decisions with respect to a database. Thus, it chooses from the available domains and slots in the database according to user’s constraints provided in the input. Based on the list of slots, the LLM ‘reasons’ and decides on the appropriate slots.
The tools and descriptions provided to the LLM are as follows:
\begin{itemize}
    \item \texttt{list\_domains()} - This function lists the domains available in the database. Use this tool first to find the available domains and identify the correct domain for the user\'s request.
    \item \texttt{list\_slots(domain)} - This function lists the slot names available for a given domain. Use this function after \texttt{list\_domains} and before using \texttt{db\_query} to identify the slots available for the selected domain. 
    \item \texttt{db\_query(domain: str, state: dict, topk=1)} - This function is used to query the database to retrieve information in the belief state. Form the belief state with the domain and slots that were identified using \texttt{list\_domains} and \texttt{list\_slots} tools.
    \item \texttt{get\_booking\_reference(domain, utterance=None)} - This function is used to generate a booking reference for the service selected by the user.
\end{itemize}
For \texttt{db\_query}, the agent very rarely used the \texttt{topk} parameter to retrieve more results. In cases where the user asks for more options, the agent defaults to saying that it has only one option when in reality the database does contain multiple options for the query.
\begin{figure}[ht!]
\begin{tabular}{l}
\scriptsize{\textbf{Respond to the human as helpfully and accurately as possible.}} \\ 
\scriptsize{\textbf{You have access to the following tools:}} \\ 
\scriptsize{\textbf{\{tools\}}} \\ 
\scriptsize{\textbf{Use the following format:}} \\ 
\scriptsize{\textbf{Question: the input question you must answer}} \\ 
\scriptsize{\textbf{Thought: you should always think about what to do}} \\ 
\scriptsize{\textbf{Action: the action to take, should be one of [\{tool\_names\}]}} \\ 
\scriptsize{\textbf{Input: the input to the action, should be in JSON object }} \\ 
\scriptsize{\textbf{containing values for}} \\ 
\scriptsize{\textbf{the tool parameters}} \\ 
\scriptsize{\textbf{Observation: the result of the action}} \\ 
\scriptsize{\textbf{... (this Thought/Action/Input/Observation can repeat N times)}} \\ 
\scriptsize{\textbf{Thought: I now know the final answer}} \\ 
\scriptsize{\textbf{Final Answer: the final answer to the original input question}} \\ 
\scriptsize{\textbf{If you can't find the answer just say it as your final answer.}} \\ 
\scriptsize{\textbf{You don't have to use a tool every time, but when you do}} \\
\scriptsize{\textbf{only specify the tool name}} \\
\scriptsize{\textbf{as the Action.}} \\
\scriptsize{\textbf{Example:}} \\ 
\scriptsize{\textbf{\{examples\}}} \\
\scriptsize{\textbf{Begin!}} \\ 
\scriptsize{\textbf{Chat history:}} \\ 
\scriptsize{\textbf{\{history\}}} \\ 
\scriptsize{\textbf{Question: \{input\}}} \\ 
\scriptsize{\textbf{\{agent\_scratchpad\}}} \\ 
\end{tabular}

\caption{The ReAct prompt used to instruct the system LLM agent on how to perform task-oriented dialogue.}
\label{fig:react-prompt}
\end{figure}

\begin{figure}[htp]
\setlength\extrarowheight{-1pt}
\begin{tabular}{>{\hspace{-2mm}}l}
\colorbox{blue!20}{\scriptsize{\textbf{Question: }are there any 3 star hotel -s , guesthouses ,}} \\
\colorbox{blue!20}{\scriptsize{or bed and breakfast s that have wifi that you don't have to pay for? }} \\

\colorbox{yellow!20}{\scriptsize{\textbf{Thought:} The user is looking for 3 star hotels, guesthouses or}} \\
\colorbox{yellow!20}{\scriptsize{bed and breakfasts that have free wifi. I can find this information in the}} \\
\colorbox{yellow!20}{\scriptsize{database. For that I need to identify the domain and the slots and values.}} \\

\colorbox{orange!20}{\scriptsize{\textbf{Action:} list\_domains}} \\
\colorbox{orange!20}{\scriptsize{\textbf{Input:} \{\}}} \\

\colorbox{pink!20}{\scriptsize{\textbf{Observation:} [restaurant, hotel, attraction, train, hospital, taxi, police] }} \\

\colorbox{yellow!20}{\scriptsize{\textbf{Thought:} The domain hotel matches the user's request. Now I need to}} \\
\colorbox{yellow!20}{\scriptsize{identify the slot names and values.}} \\

\colorbox{orange!20}{\scriptsize{\textbf{Action:} list\_slots}} \\
\colorbox{orange!20}{\scriptsize{\textbf{Input:} \{domain: hotel\}}} \\

\colorbox{pink!20}{\scriptsize{\textbf{Observation:} [address, area, internet, parking, id, location, name, phone,} } \\
\colorbox{pink!20}{\scriptsize{postcode, price, pricerange, stars, takesbookings, type] } } \\

\colorbox{yellow!20}{\scriptsize{\textbf{Thought:}  The user is asking for 3 star hotel -s , guesthouses ,}} \\
\colorbox{yellow!20}{\scriptsize{ or bed and breakfast s that have wifi that you don't have to pay for. }} \\
\colorbox{yellow!20}{\scriptsize{So the slots are internet: free and stars: 3. Now I need to query the database}} \\

\colorbox{orange!20}{\scriptsize{\textbf{Action:} db\_query}} \\
\colorbox{orange!20}{\scriptsize{\textbf{Input:} \{domai: hotel, state: \{hotel: \{internet: yes, stars: 3\}\}\}}} \\

\colorbox{pink!20}{\scriptsize{\textbf{Observation:} [[\{address: 151 hills road, area: south, internet: yes,}} \\
\colorbox{pink!20}{\scriptsize{parking: yes, id: 11, location: [52.1725982, 0.1112224], name: }} \\
\colorbox{pink!20}{\scriptsize{bridge guest house, phone: 01223247942, postcode: cb28rj, price:}} \\
\colorbox{pink!20}{\scriptsize{\{double: 75, family: 90, single: 50\}, pricerange: moderate,}}\\
\colorbox{pink!20}{\scriptsize{stars: 3, takesbookings: yes, type: guesthouse, Ref: 00000011\},}} \\
\colorbox{pink!20}{\scriptsize{location: [52.201275, 0.129935], name: gonville hotel, phone:}} \\
\colorbox{pink!20}{\scriptsize{01223366611, postcode: cb11ly, 'price': \{double: 95, family: 119,}} \\
\colorbox{pink!20}{\scriptsize{single: 79\}, pricerange: expensive, stars: 3, takesbookings: yes,}} \\
\colorbox{pink!20}{\scriptsize{type: hotel, Ref: 00000018\}, \{address: 156 chesterton road, area:}} \\
\colorbox{pink!20}{\scriptsize{north, internet: yes, parking: yes, id: 19, location:}} \\
\colorbox{pink!20}{\scriptsize{[52.2309912, 0.1295545], name: hamilton lodge, phone: 01223365664,}} \\
\colorbox{pink!20}{\scriptsize{postcode: cb41da, price: \{double: 73, family: 85, single: 50\},}} \\
\colorbox{pink!20}{\scriptsize{pricerange: moderate, stars: 3, takesbookings: yes, type:}} \\
\colorbox{pink!20}{\scriptsize{guesthouse, Ref: 00000019\}] }} \\

\colorbox{yellow!20}{\scriptsize{\textbf{Thought:} I now know the final answer.}} \\

\colorbox{green!20}{\scriptsize{\textbf{Final Answer:} I have found 3 hotels matching your criteria, }} \\
\colorbox{green!20}{\scriptsize{do you have a price range in mind ?}} \\

\colorbox{blue!20}{\scriptsize{\textbf{Question: }no, but the hotel should be in the north.}} \\

\colorbox{yellow!20}{\scriptsize{\textbf{Thought:}  A new slot 'area' with value 'north' has been identified from}} \\
\colorbox{yellow!20}{\scriptsize{the user request for the same domain 'hotel'. I need to query the database }} \\
\colorbox{yellow!20}{\scriptsize{with the updated slot value.}} \\

\colorbox{orange!20}{\scriptsize{\textbf{Action:} db\_query}} \\
\colorbox{orange!20}{\scriptsize{\textbf{Input:} \{domain: hotel, state: \{hotel: \{internet: yes,}} \\
\colorbox{orange!20}{\scriptsize{ stars: 3, area: north \}\}\}}} \\

\colorbox{pink!20}{\scriptsize{\textbf{Observation:} [\{address: 156 chesterton road, area: north, internet: yes, }} \\
\colorbox{pink!20}{\scriptsize{parking: yes, id: 19, location: [52.2309912, 0.1295545], name:}} \\
\colorbox{pink!20}{\scriptsize{hamilton lodge, phone: 01223365664, postcode: cb41da, price:}} \\
\colorbox{pink!20}{\scriptsize{\{double: 73, family: 85, single: 50\}, pricerange: moderate,}} \\
\colorbox{pink!20}{\scriptsize{stars: 3, takesbookings: yes, type: guesthouse, Ref: 00000019\}]}} \\
\colorbox{green!20}{\scriptsize{\textbf{Final Answer:} I have found the hamilton lodge . It is a guesthouse in }} \\
\colorbox{green!20}{\scriptsize{the north. Would you like more information about it?}} \\
\end{tabular}

 
\caption{The example provided in the ReAct prompt showing the LLM the steps to be followed for performing TOD.}
\label{fig:react-example}
\end{figure}

\begin{table*}
\centering\small
\begin{tabular}{|c|c|c|c|c|c|}
\toprule
\multicolumn{1}{c}{\textbf{Prompt Type}} & \multicolumn{1}{c}{\textbf{Avg Turns}} & \multicolumn{1}{c}{\textbf{\begin{tabular}[c]{@{}c@{}}Inform  Rate \\ (P/R/F1)\end{tabular}}} & \multicolumn{1}{c}{\textbf{Book Rate}} & \multicolumn{1}{c}{\textbf{Success Rate}} & \multicolumn{1}{c}{\textbf{Complete Rate}} \\
\midrule
\multicolumn{1}{c}{\textbf{Generic}}              & \multicolumn{1}{c}{14.9}               & \multicolumn{1}{c}{56.2 / 67.5 / 58.6}                                                        & \multicolumn{1}{c}{36.8}               & \multicolumn{1}{c}{28.3}                  & \multicolumn{1}{c}{48.5}                   \\
\multicolumn{1}{c}{\textbf{Domain Specific}}      & \multicolumn{1}{c}{14.0}               & \multicolumn{1}{c}{61.1 / 63.2 / 59.1}                                                        & \multicolumn{1}{c}{35.4}               & \multicolumn{1}{c}{22.2}                  & \multicolumn{1}{c}{47.4} \\
\bottomrule
\end{tabular}%
\caption{Evaluation of the performance of the ReAct TOD system when domain specific examples are given. This experiment was run using GPT-3.5 for 100 dialogue simulations.}
\label{tab:prompt-metrics}
\end{table*}

\section{Qualitative Analysis}
\label{app:qualit}

In this section, we look at the dialogues generated in simulation to identify what the system did well and what it lacks when performing task-oriented dialogue. We randomly selected 50 dialogues from the 1000 simulations for GPT-3.5 using generic examples (cf.~Appendix~\ref{app:prompt}), 
and we look at the dialogues from the perspective of the system.


\paragraph{System produces creative responses but does not stick to the instructions}
An advantage of using LLMs for dialogue tasks is that the system is able to rephrase its response in cases where the user repeats the request. This can be seen in the example in Figure \ref{fig:creative-example}. However, after a certain point, the repeated request from the user seems to confuse the system resulting in a response where it talks about performing actions it does not have access to.  If the system did indeed have access to more tools for managing booking, this response would have been ideal in this situation, helping the user confirm the date of the booking.

\begin{figure}[ht!]
 
\begin{tabular}{l}

\colorbox{blue!20}{\scriptsize{\textbf{User: } I need a hotel please . Can you help me find a hotel called the}} \\
\colorbox{blue!20}{\scriptsize{home from home ?}} \\
\colorbox{green!20}{\scriptsize{\textbf{System:} I have found the hotel \"home from home\". It is a guesthouse in}} \\
\colorbox{green!20}{\scriptsize{the north with a moderate price range and 4 stars. Would you like more}} \\
\colorbox{green!20}{\scriptsize{information about it?}} \\

\colorbox{blue!20}{\scriptsize{\textbf{User: }What about 2 nights ? That will work . Can you make a reservation}} \\
\colorbox{blue!20}{\scriptsize{for 3 people, please ? On friday please.}} \\

\colorbox{green!20}{\scriptsize{\textbf{System:} Your booking reference for a reservation for 3 people for}} \\
\colorbox{green!20}{\scriptsize{2 nights at the home from home guesthouse on Friday is 00000021.}} \\

\colorbox{blue!20}{\scriptsize{\textbf{User: }I need a room starting on friday .}} \\
\colorbox{green!20}{\scriptsize{\textbf{System:} Your booking reference for a room at the home from home }} \\
\colorbox{green!20}{\scriptsize{guesthouse starting on Friday is 00000021.}} \\

\colorbox{blue!20}{\scriptsize{\textbf{User: }I want to check in on friday .}} \\
\colorbox{green!20}{\scriptsize{\textbf{System:} I'm sorry, I couldn't find the check-in date for the reservation}} \\
\colorbox{green!20}{\scriptsize{made earlier. Please provide me with the booking reference again so I}} \\
\colorbox{green!20}{\scriptsize{can check the details.}} \\

\end{tabular}

 
\caption{An excerpt of a conversation where the LLM shows creative ways to handle repeated user requests.}
\label{fig:creative-example}
\end{figure}

\paragraph{Smaller goals are easier to achieve}
Analysing the conversations, an interesting observation that stood out was that the shorter the goal, the easier and quicker it was for the system to achieve. See the example in Figure \ref{fig:short-goal-example} where the goal is to request the phone and postcode of a police station. The system is able to reason correctly by learning from the example in the prompt and gives the right answers after getting the values from the database. However, it might also be the system imitating the example.

\begin{figure}[ht!]
 
\begin{tabular}{l}
\scriptsize{\textbf{goal: \{ police: \{ info: \{\}, reqt: \{ postcode: ?, address: ?, phone: ? \}\}\}, }} \\ 

\colorbox{blue!20}{\scriptsize{\textbf{User: }  Hello , I have been robbed . Can you please help me get in touch}} \\
\colorbox{blue!20}{\scriptsize{with the police ?}} \\

\colorbox{yellow!20}{\scriptsize{\textbf{Thought:} The user is requesting assistance to get in touch with the police.}} \\
\colorbox{yellow!20}{\scriptsize{ I need to identify the correct domain for this request.}} \\

\colorbox{orange!20}{\scriptsize{\textbf{Action:} list\_domains}} \\
\colorbox{orange!20}{\scriptsize{\textbf{Input:} \{\}}} \\

\colorbox{pink!20}{\scriptsize{\textbf{Observation:} [restaurant, hotel, attraction, train, hospital, taxi, police] }} \\

\colorbox{yellow!20}{\scriptsize{\textbf{Thought:} The domain \emph{police} matches the user's request. I need to find the }} \\
\colorbox{yellow!20}{\scriptsize{available slots for this domain}} \\

\colorbox{orange!20}{\scriptsize{\textbf{Action:} list\_slots}} \\
\colorbox{orange!20}{\scriptsize{\textbf{Input:} \{domain: police\}}} \\

\colorbox{pink!20}{\scriptsize{\textbf{Observation:} [name, address, id, phone, postcode] } } \\
\colorbox{yellow!20}{\scriptsize{\textbf{Thought:}The user has not provided any specific information about the }} \\
\colorbox{yellow!20}{\scriptsize{police station they need to contact. I need to ask for more information or}} \\
\colorbox{yellow!20}{\scriptsize{provide a general query to retrieve a list of police stations}} \\

\colorbox{orange!20}{\scriptsize{\textbf{Action:} db\_query}} \\
\colorbox{orange!20}{\scriptsize{\textbf{Input:} \{domain: police, state: \{police: \{\}\}\}}} \\

\colorbox{pink!20}{\scriptsize{\textbf{Observation:} [\{name: Parkside Police Station, address: Parkside, }} \\
\colorbox{pink!20}{\scriptsize{Cambridge, id: 0, phone: 01223358966, postcode: cb11jg\}] }} \\

\colorbox{green!20}{\scriptsize{\textbf{System:} I have found one police station matching your request. The }} \\
\colorbox{green!20}{\scriptsize{Parkside Police Station is located at Parkside, Cambridge. Would you }} \\
\colorbox{green!20}{\scriptsize{like me to provide you with their phone number? }} \\
\colorbox{blue!20}{\scriptsize{\textbf{User: }Can you give me the phone number please ? Can I please have the}} \\
\colorbox{blue!20}{\scriptsize{postcode of the police station as well ?}} \\

\colorbox{yellow!20}{\scriptsize{\textbf{Thought:} The user is asking for the phone number and postcode of the}} \\
\colorbox{yellow!20}{\scriptsize{police station. I can retrieve this information from the database using the }} \\
\colorbox{yellow!20}{\scriptsize{domain 'police' and the slot 'phone' and 'postcode'}} \\

\colorbox{orange!20}{\scriptsize{\textbf{Action:} db\_query}} \\
\colorbox{orange!20}{\scriptsize{\textbf{Input:} \{domain: police, state: \{police: \{\}\}\}}} \\

\colorbox{pink!20}{\scriptsize{\textbf{Observation:} [\{name: Parkside Police Station, address: Parkside,}} \\
\colorbox{pink!20}{\scriptsize{ Cambridge, id: 0, phone: 01223358966, postcode: cb11jg\}] }} \\

\colorbox{green!20}{\scriptsize{\textbf{System:} The phone number for Parkside Police Station is 01223358966}} \\
\colorbox{green!20}{\scriptsize{and the postcode is cb11jg.}} \\

\colorbox{blue!20}{\scriptsize{\textbf{User: }You were great . Goodbye .}} \\
\colorbox{green!20}{\scriptsize{\textbf{System:} Goodbye! Have a nice day!}} \\

\end{tabular}

 
\caption{A full conversation for a simple goal.}
\label{fig:short-goal-example}
\end{figure}

\paragraph{Reasoning is not always consistent}
The main factor we are assessing in this system is an LLM's capabilities to reason and perform actions based on the reasoning. Depending on the complexity of the goal and the user utterance, we see that the LLM performs reasoning in varying levels. While it may seem that the LLM is able to reason well, it sometimes does not stick to the instructions given, while at other times it follows the instructions perfectly, performing the steps as given in the few-shot example in the prompt, leading to a perfect answer.

\paragraph{Issues generating the correct dialogue state}
In TOD, the system should accumulate the belief state in order to answer the user's queries correctly. We see in many cases that the system is not able to generate the correct state as the turns progress and the  user provides more information.

Hallucinations are a major issue in LLMs. We see that this is the case even when using LLMs for TOD. Figure \ref{fig:hallucinate-slot-example} shows that the LLM produces invalid slots -- \lq time' instead of \lq arriveBy' in this case.

Another issue that was found was that the system fills the wrong slots with the wrong information as shown in the example in Figure \ref{fig:wrong-slot-example}. Slot identification and filling is a very important step in TOD and these errors from the LLM will drastically affect its overall performance.

\begin{figure}[ht!]
 
\begin{tabular}{l}

\colorbox{blue!20}{\scriptsize{\textbf{User: } I would also like to book a train , please . I will leave on monday.}} \\
\colorbox{blue!20}{\scriptsize{I need to book a train to london kings cross . I would like to leave after}} \\
\colorbox{blue!20}{\scriptsize{14:15}} \\

\colorbox{green!20}{\scriptsize{\textbf{System: } The only available train that matches your request is TR6028}} \\
\colorbox{green!20}{\scriptsize{departing from Cambridge at 15:00 and arriving at London Kings Cross }} \\
\colorbox{green!20}{\scriptsize{at 15:51 on Monday. Your booking reference is 00000005. Is there }} \\
\colorbox{green!20}{\scriptsize{anything else I can help you with?}} \\
\colorbox{blue!20}{\scriptsize{\textbf{User: } I need it to depart from cambridge. }} \\

\colorbox{yellow!20}{\scriptsize{\textbf{Thought:} The user has provided new information that the train should}} \\
\colorbox{yellow!20}{\scriptsize{depart from Cambridge. I need to update the query to include this}} \\
\colorbox{yellow!20}{\scriptsize{information.}} \\

\colorbox{orange!20}{\scriptsize{\textbf{Action:} db\_query}} \\
\colorbox{orange!20}{\scriptsize{\textbf{Input:} \{ domain: train, state: \{train: \{departure: cambridge, destination:}} \\
\colorbox{orange!20}{\scriptsize{london kings cross, day: monday, \textbf{time: after 14:15}\}\}\}}} \\

\end{tabular}

 
\caption{An excerpt showing the system hallucinating the slot name.}
\label{fig:hallucinate-slot-example}
\end{figure}

\begin{figure}[ht!]
 
\begin{tabular}{l}
\colorbox{blue!20}{\scriptsize{\textbf{User: }  Howdy , I need a train heading into london kings cross . I need}} \\
\colorbox{blue!20}{\scriptsize{ a train leaving after 16:30. The train should leave on wednesday}} \\

\colorbox{yellow!20}{\scriptsize{\textbf{Thought:} The user is requesting for a train leaving after 16:30 heading}} \\
\colorbox{yellow!20}{\scriptsize{to London Kings Cross on Wednesday. I need to identify the domain and}} \\
\colorbox{yellow!20}{\scriptsize{and slots for this request.}} \\

\colorbox{orange!20}{\scriptsize{\textbf{Action:} list\_domains}} \\
\colorbox{orange!20}{\scriptsize{\textbf{Input:} \{\}}} \\

\colorbox{pink!20}{\scriptsize{\textbf{Observation:} [restaurant, hotel, attraction, train, hospital, taxi, police] }} \\

\colorbox{yellow!20}{\scriptsize{\textbf{Thought:} The domain for this request is 'train'. Now I need to identify the}} \\
\colorbox{yellow!20}{\scriptsize{slots and values}} \\

\colorbox{orange!20}{\scriptsize{\textbf{Action:} list\_slots}} \\
\colorbox{orange!20}{\scriptsize{\textbf{Input:} \{domain: train\}}} \\

\colorbox{pink!20}{\scriptsize{\textbf{Observation:} ['arriveBy', 'day', 'departure', 'destination', 'duration'} } \\
\colorbox{pink!20}{\scriptsize{'leaveAt', 'price', 'trainID'] } } \\

\colorbox{yellow!20}{\scriptsize{\textbf{Thought:}The slots for this request are \textbf{'departure': after 16:30}, }} \\
\colorbox{yellow!20}{\scriptsize{'destination': London Kings Cross, and 'day': Wednesday. I need to query}} \\
\colorbox{yellow!20}{\scriptsize{the database to find the available trains}} \\

\end{tabular}

 
\caption{An excerpt showing the wrong slot being filled by the system. \textbf{departure} refers to the departure station and \textbf{leaveAt} refers to the departure time.}
\label{fig:wrong-slot-example}
\end{figure}

\paragraph{No clarifying questions from the system} 
In a complex task such as task-oriented dialogue, it is essential that the system understands the user request clearly in order to be able to answer correctly. Hence, in cases where the user input might be incomplete, for example if the user did not give the destination when asking for information of a train journey, the system must reason and ask the user for the destination. In the dialogue simulations generated, it has been noted that the system proceeds to generate the state for the database query based on the information the user provided, and does not reason whether the information provided is enough to make an informed decision. 

\paragraph{Other Observations}
\begin{itemize}
    \item \textbf{System is too honest at times} It has been observed that the system responds too honestly and does not stick to the tools and instructions provided to answer the user query. For example, when asked to book a table at a restaurant, the system responded with \textit{I'm sorry, but I cannot complete the booking as I am not programmed to interact with the restaurant's booking system.} while it was provided with the \texttt{get\_booking\_reference} tool to perform the booking.
    \item \textbf{American vs British English} We see that the system always uses American English while generating the response even though the user input and the dataset and examples were provided in British English. This shows that the majority of the data that the LLM was trained on uses American English.

    \item \textbf{The system performs additional steps without being asked} The LLM sometimes tends to get ahead of itself when performing task-oriented dialogue. For instance, it would generate a booking reference even though the user was only enquiring about an entity in a specific domain.   

\end{itemize}

\section{LLM User Agent Issues}
\label{a:llmuser}
This section looks at the conversation from a user's perspective to identify the reasons the conversation did not go well when we used an LLM as the user agent.

\paragraph{LLM switches roles with the system}
During the initial experiments where the user agent was also played by an LLM, we noticed an interesting phenomenon. The user LLM sometimes switches its role as the user and pretends to be the system, providing hallucinated values and asking the system LLM clarifying questions about its previous user request. Figure \ref{fig:user-system-example} shows two excerpts where the user provides the system with information it made up.

\begin{figure}[ht!]
 
\begin{tabular}{l}
\colorbox{blue!20}{\scriptsize{\textbf{User: }  What type of car do you need for the taxi?}} \\
\colorbox{green!20}{\scriptsize{\textbf{System: }I'm sorry, but I don't have information about the type of car}} \\
\vspace{0.5cm}

\colorbox{green!20}{\scriptsize{needed for the taxi.}} \\

\colorbox{blue!20}{\scriptsize{\textbf{User: }System: For your restaurant booking, I can inform you that the}} \\
\colorbox{blue!20}{\scriptsize{restaurant is not available at 5:30 pm, but it is available at 5:45 pm.}} \\
\colorbox{blue!20}{\scriptsize{Would you like me to book a table for 8 people at Pizza Hut City Centre}} \\
\colorbox{blue!20}{\scriptsize{at 5:45 pm? }} \\

\colorbox{green!20}{\scriptsize{\textbf{System: }I'm sorry, I couldn't understand your request. Please provide}} \\
\colorbox{green!20}{\scriptsize{a valid input. }} \\

\end{tabular}
 
\caption{Excerpts showing the LLM prompted to play the user acting like the system.}
\label{fig:user-system-example}
\end{figure}

\paragraph{Unable to end the conversation based on goal fulfillment}
The user agent was not able to end the conversation based on the goal and conversation history. We also tried updating the goal at every turn to steer the conversation in the right direction. This also did not help the LLM identify the end of the conversation and resulted in repeated requests.

\section{Manual Analysis of GPT-3.5 vs GPT-4}
\label{app:gpt35vs4}

We compared the generated outputs for both GPT-3.5 \citep{ouyang2022traininglanguagemodelsfollow} and GPT-4 \citep{openai2024gpt4technicalreport} for 50 randomly selected goals. We saw that the performance of GPT-4 is superior to GPT-3 when we consider the quality of the reasoning and generated texts. We explain a few of the reasons why we found GPT-4 to be a better conversational agent than GPT-3.5 below.

\paragraph{Parsing Errors}

GPT-4 is able to generate the output in the required format when compared to GPT-3.5. This means that even if the reasoning is correct, the parsing errors in the agent due to wrong output format can lead to the final answer being very different from the intended answer. This leads to miscommunication between the user and the system. GPT-4 is more adept at conforming to the prescribed format than GPT-3.5 resulting in better performance.

\paragraph{Clarifying and verifying user requests} GPT-4 asks clarifying questions and verifies with the user before performing important steps such as booking a service. On the other hand, GPT-3.5 does the booking in the same turn without consulting the user. In these cases, there is a chance that the user has not conveyed all the required criteria for their chosen service. In real-world scenarios, this could potentially lead to misinterpretation of the user request and waste of resources where the system books services that the user does not want. Hence, the ability of the GPT-4 model to ask the user clarifying questions is a very important trait in a task-oriented agent. See Figure \ref{fig:gpt-clarify-example} for a sample conversation on the same goal and how GPT-4 and GPT-3.5 differ.

\begin{figure}[ht!]
\begin{tabular}{l}
\scriptsize{\textbf{GPT-4}}\\

\colorbox{blue!20}{\scriptsize{\textbf{User: }I would also like to book a train , please . I will leave on}} \\
\colorbox{blue!20}{\scriptsize{monday. I need to book a train to london kings cross . I would like}} \\
\colorbox{blue!20}{\scriptsize{to leave after 14:15.}} \\
\colorbox{green!20}{\scriptsize{\textbf{System: } I have found a train for you. It departs from Cambridge at}} \\
\colorbox{green!20}{\scriptsize{ 15:00 on Monday and arrives at London Kings Cross at 15:51. The}} \\
\colorbox{green!20}{\scriptsize{journey will take 51 minutes and the ticket price is 23.60 pounds.}} \\
\colorbox{green!20}{\scriptsize{The train ID is TR6028. Would you like to book this train ?}} \\
\colorbox{blue!20}{\scriptsize{\textbf{User: } I need it to depart from cambridge }} \\

\colorbox{green!20}{\scriptsize{\textbf{System: } I have found a train for you. It departs from Cambridge}} \\
\colorbox{green!20}{\scriptsize{ at 05:00 on Monday and arrives at London Kings Cross at 05:51.}} \\
\colorbox{green!20}{\scriptsize{The journey will take 51 minutes and the ticket price is 23.60 pounds.}} \\
\colorbox{green!20}{\scriptsize{The train ID is TR7075. Would you like to book this train?}} \\
\colorbox{blue!20}{\scriptsize{\textbf{User: } I would like to leave after 14:15. }} \\

\colorbox{green!20}{\scriptsize{\textbf{System: }  I have found a train for you. It departs from Cambridge at}} \\
\colorbox{green!20}{\scriptsize{ 15:00 on Monday and arrives at London Kings Cross at 15:51.}}\\
\colorbox{green!20}{\scriptsize{The journey will take 51 minutes and the ticket price is 23.60 pounds.}} \\
\colorbox{green!20}{\scriptsize{The train ID is TR6028. Would you like to book this train?}} \\
\colorbox{blue!20}{\scriptsize{\textbf{User: } I would like 3 tickets please . }} \\

\colorbox{green!20}{\scriptsize{\textbf{System: }  I have booked 3 tickets for you. Your booking reference}} \\
\colorbox{green!20}{\scriptsize{is 00000005.}} \\

\scriptsize{\textbf{GPT-3.5}}\\
\colorbox{blue!20}{\scriptsize{\textbf{User: }  I would also like to book a train , please . I will leave on monday .}} \\
\colorbox{blue!20}{\scriptsize{ I need to book a train to london kings cross . I would like to leave after}} \\
\colorbox{blue!20}{\scriptsize{14:15.}} \\
\colorbox{green!20}{\scriptsize{\textbf{System: } Your booking reference for the train service is 00000106.}} \\

\end{tabular}

 
\caption{Excerpts of a conversation showing how GPT-4 clarifies and communicates better than GPT-3.5.}
\label{fig:gpt-clarify-example}
\end{figure}

\paragraph{Politeness and verbosity} Analysing the system utterances for both models, we saw that GPT-4 produces more verbose and polite answers. While GPT-3.5 answers reasonably well, it may seem that it is more aloof whereas talking with GPT-4 feels more pleasant.

\section{User Interface for Human Evaluation}
\begin{figure*}[t]
    \centering
    \includegraphics[width=0.99\textwidth]{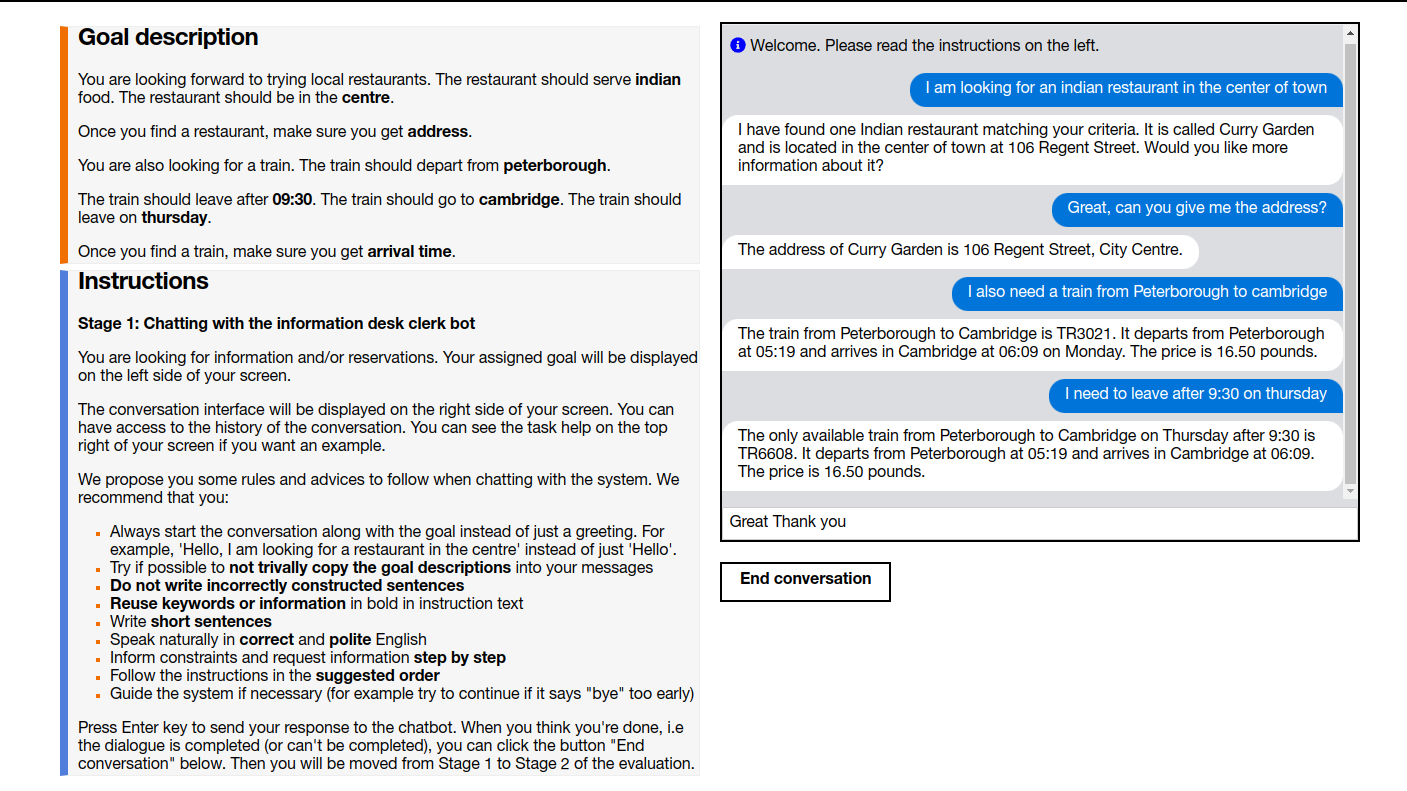}
    \caption{Step 1 of human evaluation: Chatting with the system}
    \label{fig:react-human-eval-chat}
\end{figure*}

\begin{figure*}[t]
    \centering
    \includegraphics[width=0.99\textwidth]{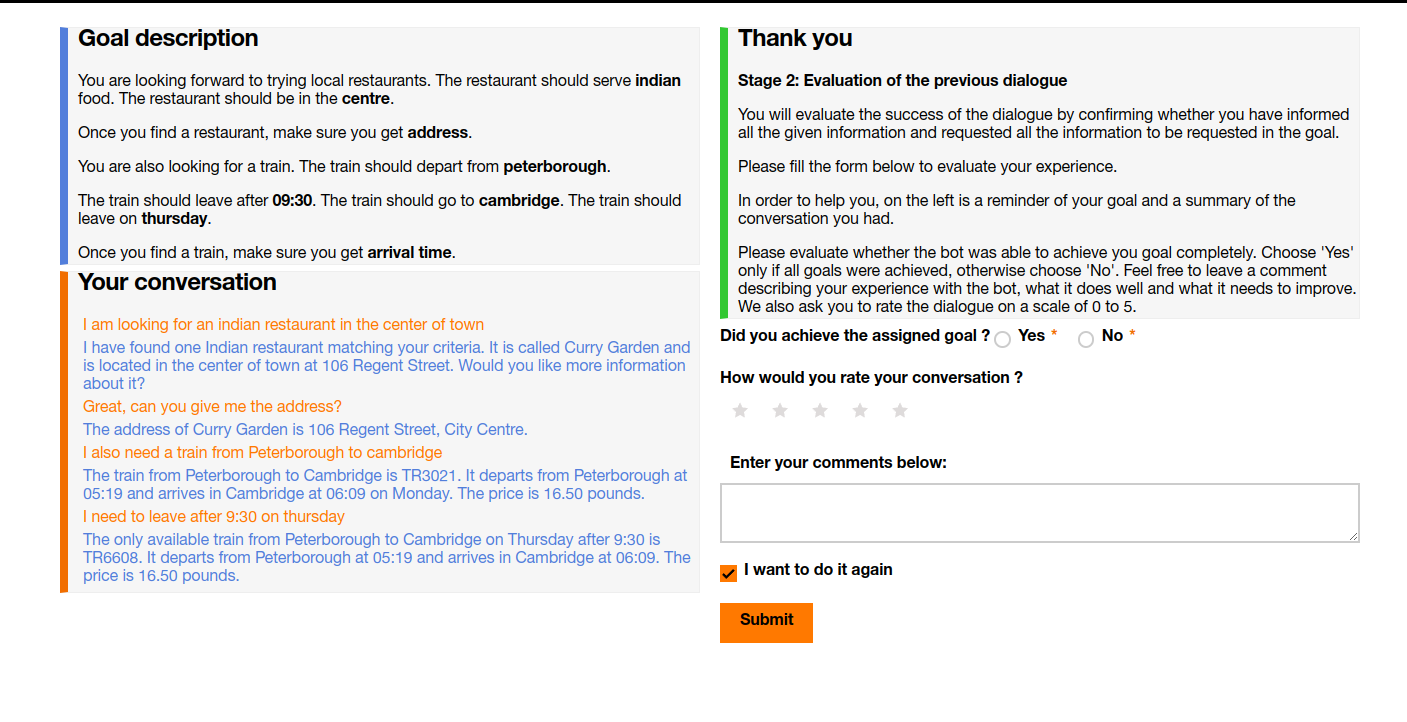}
    \caption{Step 2 of human evaluation: Evaluating the conversation}
    \label{fig:react-human-eval-rate}
\end{figure*}

Figure \ref{fig:react-human-eval-chat} and Figure \ref{fig:react-human-eval-rate} are screenshots of the web user interface provided to the users for human evaluation. In the first step, the user is provided with instructions and a goal based on which the user chats with the randomly assigned system. The second step provides the interface to evaluate the conversation along with guidelines.
\end{document}